\documentclass[conference]{IEEEtran}
\IEEEoverridecommandlockouts

\usepackage{graphicx}   
\usepackage{cite}       
\usepackage{amsmath,amssymb}
\usepackage{hyperref}
\usepackage{xurl}
\usepackage{array}
\usepackage{caption}
\usepackage{algpseudocode}

\usepackage{titlesec}
\titlespacing*{\subsection}{0pt}{1ex}{1ex}

\usepackage[dvipsnames]{xcolor}
\usepackage{multirow}
\usepackage{pgfplots}
\pgfplotsset{compat=1.18}

\usepackage[most]{tcolorbox}
\usepackage{newfloat}
\usepackage{float}
\usepackage{enumitem}

\begin{document}

\title{\vspace*{5mm}CADSmith: Multi-Agent CAD Generation \\with Programmatic Geometric Validation}

\author{Jesse Barkley$^{1}$, Rumi Loghmani$^{1}$, and Amir Barati Farimani$^{1}$%
\thanks{\makeatletter\footnotesize%
$^{1}$With the Department of Mechanical Engineering, Carnegie Mellon University,%
{\ttfamily\{jabarkle, rloghman, afariman\}@andrew.cmu.edu}
\makeatother}
}

\maketitle

\begin{abstract}
Existing methods for text-to-CAD generation either operate in a single pass with no geometric verification or rely on lossy visual feedback that cannot resolve dimensional errors. We present CADSmith, a multi-agent pipeline that generates CadQuery code from natural language. It then undergoes an iterative refinement process through two nested correction loops: an inner loop that resolves execution errors and an outer loop grounded in programmatic geometric validation. The outer loop combines exact measurements from the OpenCASCADE kernel (bounding box dimensions, volume, solid validity) with holistic visual assessment from an independent vision-language model Judge. This provides both the numerical precision and the high-level shape awareness needed to converge on the correct geometry. The system uses retrieval-augmented generation over API documentation rather than fine-tuning, maintaining a current database as the underlying CAD library evolves. We evaluate on a custom benchmark of 100 prompts in three difficulty tiers (T1 through T3) with three ablation configurations. Against a zero-shot baseline, CADSmith achieves a 100\% execution rate (up from 95\%), improves the median F1 score from 0.9707 to 0.9846, the median IoU from 0.8085 to 0.9629, and reduces the mean Chamfer Distance from 28.37 to 0.74, demonstrating that closed-loop refinement with programmatic geometric feedback substantially improves the quality and reliability of LLM-generated CAD models.
\end{abstract}


\begin{figure}[t]
    \centering
    \includegraphics[width=0.95\columnwidth]{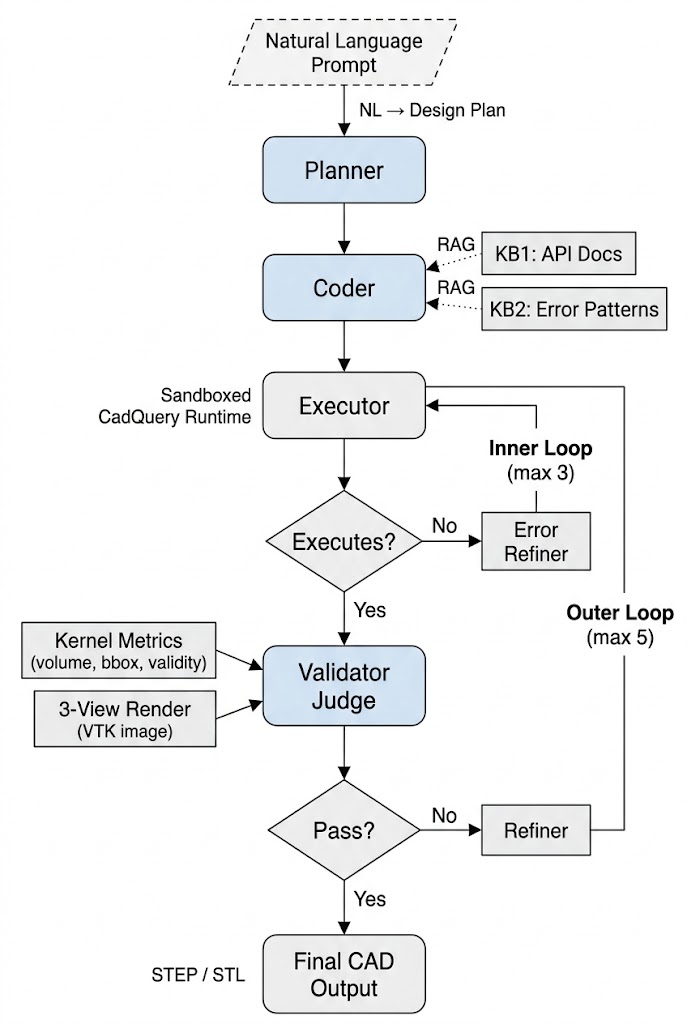}
    \caption{CADSmith pipeline overview. A natural language prompt flows through five agents with two nested correction loops: an inner loop for execution errors and an outer loop for geometric refinement driven by kernel metrics and three-view vision.}
    \label{fig:overview}
\end{figure}

\section{Introduction}

Computer-aided design (CAD) remains one of the most expertise-intensive tasks in engineering and fabrication. Creating even moderately complex parts requires fluency with specialized software, deep knowledge of geometric constraints, and careful attention to dimensional precision \cite{kasik2005ten}. As artificial intelligence continues to transform adjacent fields, from image generation to code synthesis, CAD modeling has remained largely manual \cite{hunde2022future, heidari2024geometric}. The core difficulty is that CAD outputs are not merely visual; they must be dimensionally exact, geometrically valid, and suitable for downstream manufacturing. A generated part that \textit{looks} correct but has a bounding box error of even a few millimeters, or a topology with missing faces, is unusable in practice.

This fundamental requirement for precision creates a need for quantifiable metric comparison during large language model (LLM) CAD generation. LLMs can produce plausible CAD code from natural language descriptions, but they are known to routinely hallucinate dimensions and misuse APIs \cite{ji2023hallucination, naveed2023llm_overview}. Correcting these errors requires exact geometric feedback: the actual bounding box length versus the target, the number of faces produced versus expected, the measured volume versus the design intent. The model will have little to act on if there is ungrounded visual feedback alone and no numerical measurements. However, even with precise dimensional feedback, there is a complementary need to periodically visualize the shape at a high level, much as a human engineer steps back from a workbench to inspect the overall form. Another challenge is that while geometric measurements can confirm that a part's dimensions are converging, it may not detect that the shape is fundamentally malformed or trending in an unintended direction. Without both forms of feedback working synchronously, a system risks producing parts that satisfy numerical tolerances yet bear little resemblance to the original prompt, or conversely, parts that look roughly correct but fail basic dimensional checks.

Two recent works have made significant progress on text-to-CAD generation through code synthesis. Query2CAD \cite{badagabettu2024query2cad} pairs GPT-4 with FreeCAD \cite{freecad}, using BLIP-2 \cite{li2023blip2} to provide visual feedback for iterative refinement. While this demonstrated the viability of closed-loop CAD generation, the encoded visual representations lose the geometric precision needed to resolve dimensional errors. Text-to-CadQuery \cite{xie2025texttocadquery} takes a different approach, fine-tuning models on 170,000 description-code pairs to generate CadQuery \cite{cadquery} scripts directly. This produced strong results on mesh-similarity metrics such as Chamfer Distance and F1 score, but the system operates in a single pass with no iterative correction and no validation that the output geometry matches the input specification. Neither system programmatically verifies that the generated part satisfies the dimensional and topological requirements of the original prompt.

The broader literature on LLM self-correction suggests that iterative refinement with structured feedback can substantially improve output quality \cite{madaan2023selfrefine, pan2023autocorrecting}. Yet applying this paradigm to CAD generation requires a feedback signal that is both precise enough for an LLM to act on and comprehensive enough to catch failures that purely numerical checks miss. Image-based evaluation metrics such as VQAScore \cite{lin2024vqascore} capture high-level visual similarity but cannot resolve millimeter-scale dimensional errors. Conversely, kernel-level geometric measurements are exact but blind to global shape coherence. Recent work has shown that vision-language models can serve as human-aligned evaluators for 3D generation tasks \cite{wu2024gpt4v_evaluator, zhang2023gpt4v_generalist}, but no prior CAD system has combined this capability with programmatic geometric validation in a closed refinement loop.

We present CADSmith, a multi-agent pipeline for text-to-CAD generation that addresses these gaps. Following the principle of hierarchical task decomposition \cite{simon1962architecture}, CADSmith decomposes the generation problem across specialized agents: a Planner that converts natural language into a structured design specification, a Coder that generates CadQuery code informed by retrieval-augmented generation (RAG) over API documentation, a sandboxed Executor, and a Validator that combines OpenCASCADE (OCCT) kernel checks with a vision-language model Judge. When validation fails, a Refiner agent receives both the exact geometric discrepancies and the Judge's visual assessment, subsequently producing a corrected code for re-execution. This creates two nested correction loops: an inner loop for execution errors and an outer loop for geometric refinement. By grounding refinement in programmatic measurements rather than relying on fine-tuning, CADSmith remains current as the CadQuery API evolves \cite{hu2022lora, brown2020language}, and by using a stronger model (Claude Opus \cite{anthropic2024claude3}) as the Judge rather than the same model that generated the code, the system avoids the self-confirmation bias inherent in single-model refinement. Fig.~\ref{fig:overview} shows the full pipeline of our agentic system. 

\subsection{Contributions}
In summary, our contributions are as follows:

\begin{itemize}
    \item \textbf{Multi-agent architecture.} A decomposed pipeline of specialized agents (Planner, Coder, Executor, Validator, Refiner) for text-to-CAD generation, replacing the monolithic single-model approaches used in prior work.
    \item \textbf{Dual-loop programmatic geometric validation.} An inner loop that corrects execution errors and an outer loop that uses OpenCASCADE kernel measurements (bounding box dimensions, volume, face/edge/vertex counts, solid validity) to drive iterative geometric correction.
    \item \textbf{Independent VLM-as-Judge evaluation.} A vision-language model Judge (Claude Opus) that operates independently of the generation agents (Claude Sonnet), combining three-view rendered inspection with kernel metrics to avoid self-confirmation bias.
    \item \textbf{RAG over API documentation.} A retrieval-augmented generation approach over CadQuery API documentation that eliminates the need for fine-tuning, allowing the system to remain current as the underlying CAD library evolves.
\end{itemize}

\section{Related Work}

The rapid advancement of large language models has created new possibilities for automating engineering tasks that were previously considered too structured or too precise for generative approaches \cite{brown2020language, naveed2023llm_overview}. Models such as GPT-4, Claude \cite{anthropic2024claude3}, and open-weight alternatives have demonstrated strong code generation capabilities, while techniques such as low-rank adaptation \cite{hu2022lora} have made domain-specific fine-tuning practical. However, LLMs remain prone to hallucination \cite{ji2023hallucination}, particularly when generating outputs that must satisfy exact numerical constraints. This tension between generative fluency and dimensional precision is central to the challenge of applying LLMs to CAD generation.

\subsection{Text-to-3D and Text-to-CAD Generation}

Generating three-dimensional shapes from natural language has been approached from multiple directions. Surveys by Xu et al. \cite{xu2023dl3dsurvey} and Heidari and Iosifidis \cite{heidari2024geometric} provide broad overviews of deep learning methods for 3D shape generation and CAD representation learning, respectively. Early learned approaches such as DeepCAD \cite{wu2021deepcad} trained generative models directly on CAD command sequences, producing parametric models but requiring large supervised datasets. Text2CAD \cite{khan2024text2cad} extended this direction by pairing natural language prompts with sequential CAD operations at varying levels of detail.

More recently, LLM-based approaches have emerged as an alternative to learned sequence models. LLM4CAD \cite{li2025llm4cad, sun2025llm4cad_finetuned} uses multimodal large language models to generate CAD construction sequences, demonstrating that pre-trained language models can be adapted for structured geometric output. OpenECAD \cite{yuan2024openecad} applies vision-language models to produce editable CAD designs, while CAD-MLLM \cite{xu2024cadmllm} unifies multiple input modalities (text, images, point clouds) for CAD generation through a single multimodal model. In the broader design space, Ma et al. \cite{ma2023conceptual} explored LLMs for conceptual design generation, Edwards et al. \cite{edwards2024sketch2prototype} combined generative AI with rapid prototyping workflows, and Liu et al. \cite{liu20233dalle} integrated text-to-image models into 3D design pipelines. Li et al. \cite{li2024llm_manufacturing} survey the growing role of LLMs across manufacturing more broadly. Commercial tools such as Zoo \cite{zoo_textocad} have also begun offering text-to-CAD capabilities, though without published methods or reproducible benchmarks.

The two systems most closely related to CADSmith both generate code rather than learned representations. Query2CAD \cite{badagabettu2024query2cad} generates FreeCAD \cite{freecad} Python scripts using GPT-4, using BLIP-2 \cite{li2023blip2} for visual feedback in a self-refinement loop. Text-to-CadQuery \cite{xie2025texttocadquery} fine-tunes models on 170,000 description-code pairs to generate CadQuery \cite{cadquery} scripts, evaluating with Chamfer Distance, F1, and IoU against reference meshes derived from DeepCAD \cite{wu2021deepcad} and Text2CAD \cite{khan2024text2cad} datasets. Machado et al. \cite{machado2019parametric} compare programmatic CAD approaches (OpenSCAD, FreeCAD Python), providing context for why CadQuery's clean Python API is well suited to LLM-based generation. CADSmith builds on the code generation paradigm but adds what both prior systems lack: closed-loop geometric validation that programmatically verifies dimensional and topological correctness before accepting a result.

\subsection{LLM Self-Refinement and Evaluation}

The idea of using LLMs to iteratively refine their own outputs has gained significant traction. Madaan et al. \cite{madaan2023selfrefine} introduced Self-Refine, demonstrating that a single model can improve its outputs through successive rounds of self-generated feedback. Pan et al. \cite{pan2023autocorrecting} survey the broader landscape of LLM self-correction strategies, identifying feedback quality as the critical bottleneck: refinement only helps when the feedback signal is informative enough to guide correction.

In the context of 3D generation, evaluation itself presents a challenge. Wu et al. \cite{wu2024gpt4v_evaluator} showed that GPT-4V can serve as a human-aligned evaluator for text-to-3D tasks, and Zhang et al. \cite{zhang2023gpt4v_generalist} demonstrated its effectiveness as a generalist vision-language evaluator across diverse tasks. These findings support the use of vision-language models as judges, but prior applications have focused on evaluation as a final scoring step rather than as feedback within a refinement loop. Image-based metrics such as VQAScore \cite{lin2024vqascore} capture semantic similarity between generated shapes and text prompts but cannot resolve the millimeter-scale dimensional errors that matter in engineering. Query2CAD's use of BLIP-2 \cite{li2023blip2} for visual feedback demonstrated closed-loop refinement, but the encoded representations lose the geometric precision needed to guide corrections. Text-to-CadQuery used Gemini 2.0 Flash \cite{gemini_api} for visual evaluation but only as a post-hoc assessment, not within a refinement loop.

CADSmith addresses this gap by combining programmatic kernel measurements from OpenCASCADE with a vision-language model Judge (Claude Opus) in a closed refinement loop. The kernel metrics provide exact dimensional feedback that an LLM can act on, while the Judge provides holistic shape assessment that catches failures the metrics alone would miss. Using a stronger model as the Judge than the model that generated the code avoids the self-confirmation bias identified in single-model refinement approaches \cite{pan2023autocorrecting}.

\section{Methods}

CADSmith generates CAD models from natural language through a pipeline of five specialized agents: Planner, Coder, Executor, Validator, and Refiner. These agents operate within two nested correction loops. An inner loop handles code execution errors: when the Executor fails, an Error Refiner agent diagnoses the traceback with retrieval-augmented context and produces corrected code, retrying up to three times. An outer loop handles geometric errors: when code executes but produces incorrect geometry, the Validator provides structured feedback and the Refiner modifies the code, iterating up to five times. The following subsections describe the benchmark, each pipeline stage, and the evaluation metrics.

\subsection{Benchmark Dataset}

We constructed a benchmark of 100 natural language prompts paired with hand-written CadQuery reference scripts across three difficulty tiers:

\begin{itemize}[nosep,leftmargin=*]
    \item \textbf{T1: Basic Primitives} (50 entries). Single geometric shapes: boxes, cylinders, cones, tori, prisms, domes. One to three CadQuery operations each.
    \item \textbf{T2: Engineering Parts} (25 entries). Parts requiring boolean operations: brackets, flanges, gears, shafts, plates with hole patterns, counterbored fasteners. Three to eight operations each.
    \item \textbf{T3: Complex Parts} (25 entries). Multi-feature parts requiring workplane changes, lofts, sweeps, shells, revolves, and multi-body unions. Five to fifteen operations each.
\end{itemize}

\begin{figure}[t]
    \centering
    \includegraphics[width=0.95\columnwidth]{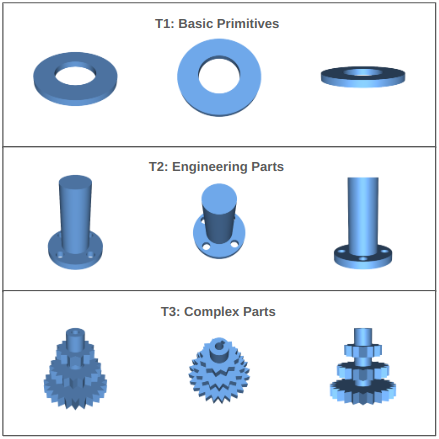}
    \caption{Representative benchmark parts across three difficulty tiers. T1 parts are single primitives, T2 parts involve boolean combinations and hole patterns, and T3 parts require multi-step construction with workplane changes, sweeps, and complex feature interactions.}
    \label{fig:tiers}
\end{figure}

Every reference script was written by hand, executed to confirm it produces a valid solid, and visually inspected before inclusion. This was motivated by quality issues observed in existing LLM-generated datasets, where reference geometry silently diverged from prompt intent through wrong dimensions, misplaced features, or simplified geometry. All prompts follow a consistent convention: explicit millimeter dimensions, axis-annotated orientations (XY base plane, Z-up), origin-centered geometry, and coordinate positions for features such as bolt holes and slots. Fig.~\ref{fig:tiers} shows representative parts from each tier.

\subsection{Planner Agent}

The Planner agent converts a natural language prompt into a structured design plan. It receives the raw user prompt and outputs a JSON object containing: a component list with sub-part descriptions, target bounding box dimensions in millimeters, geometric constraints (hole counts, hole diameters, symmetry properties), and notes for downstream agents. The Planner does not generate CadQuery code. Its role is to decompose the design intent into an unambiguous specification that the Coder can implement.

An example input prompt is shown in Fig.~\ref{fig:prompt_example}. The Planner would extract from this prompt: three cylindrical components with explicit Z-ranges, a 14\,mm center bore, six bolt holes on a 38\,mm pitch circle, and a 5\,mm keyway slot.

\begin{figure}[t]
\begin{tcolorbox}[
    colback=gray!5,
    colframe=gray!60,
    fonttitle=\bfseries\small,
    title={Example T3 Prompt (Flanged Shaft Coupling)},
    boxrule=0.4pt,
    arc=2pt,
    left=4pt, right=4pt, top=2pt, bottom=2pt,
    fontupper=\small
]
\textit{``A flanged shaft coupling standing along the Z axis, centered at the origin in X and Y. The coupling consists of three coaxial cylindrical sections: a bottom flange disc 50\,mm in diameter and 10\,mm thick (Z=0 to Z=10), a central hub cylinder 28\,mm in diameter and 40\,mm long (Z=10 to Z=50), and a top flange disc 50\,mm in diameter and 10\,mm thick (Z=50 to Z=60). Six bolt holes 6.5\,mm in diameter are equally spaced on a 38\,mm pitch circle diameter. A 14\,mm diameter center bore runs the full 60\,mm length. A rectangular keyway slot 5\,mm wide and 2.5\,mm deep is cut along the bore wall on the +X side.''}
\end{tcolorbox}
\caption*{}
\vspace{-12pt}
\captionof{figure}{Example benchmark prompt from T3 (Complex Parts). This prompt specifies the part shown in Fig.~\ref{fig:vision_views}. The Planner agent parses this into a structured design plan with explicit dimensions, component decomposition, and feature specifications.}
\label{fig:prompt_example}
\end{figure}

\subsection{Coder Agent and Retrieval-Augmented Generation}

The Coder agent receives the design plan and produces an executable CadQuery Python script. Rather than relying on fine-tuning, the Coder is augmented with two retrieval knowledge bases at inference time:

\textbf{KB1: API Documentation.} A curated reference of 155 CadQuery Workplane method entries covering primitives, sketches, extrusions, booleans, fillets, sweeps, lofts, holes, transforms, selections, and arrays. Each entry includes the method signature, description, usage example, and known pitfalls. The knowledge base also includes a selectors reference (conditionally injected when the prompt involves face or edge selection) and 28 worked code examples demonstrating common construction patterns. Keywords extracted from the design plan and prompt are matched against method categories to retrieve relevant entries.

\textbf{KB2: Error-Solution Patterns.} A database of 25 patterns covering common CadQuery and OpenCASCADE failure modes: fillet radius violations, boolean operation errors, wire closure failures, arc construction issues, extrusion crashes, and selector misuse. Each pattern contains trigger keywords, a root cause explanation, and fix instructions with before/after code. When the Error Refiner receives a traceback, keywords are matched to retrieve relevant patterns.

Both knowledge bases use keyword matching rather than embedding-based retrieval. At this corpus scale, keyword matching is deterministic, requires no additional dependencies, and avoids the overhead of maintaining an embedding model. The retrieval interface is designed so the backend can be replaced with vector search if the corpus grows.

The Coder and Error Refiner agents both receive KB1 context. The Error Refiner additionally receives KB2 context matched against the specific error traceback.

\subsection{Executor}

The Executor runs generated CadQuery scripts in an isolated Python subprocess with a 60-second timeout. On success, it exports the solid to STEP and STL formats and extracts geometric measurements directly from the OpenCASCADE kernel: volume, bounding box dimensions, center of mass, face/edge/vertex counts, and solid validity. On failure, it captures the full error traceback and error type. The Executor is deterministic and involves no LLM calls.

If execution fails, the inner correction loop activates: the Error Refiner agent receives the failing code, the traceback, and retrieval context from both KB1 and KB2, then produces corrected code. This repeats up to three times before declaring a code-execution failure.

\subsection{Geometric Validation (Outer Loop)}

When code executes successfully, the Validator determines whether the resulting geometry satisfies the original prompt. This is implemented as an LLM-as-a-Judge architecture using Claude Opus \cite{anthropic2024claude3}, a stronger model than the Claude Sonnet used for code generation, to avoid confirmation bias from self-evaluation \cite{pan2023autocorrecting}.

\begin{figure*}[t]
    \centering
    \includegraphics[width=0.85\textwidth]{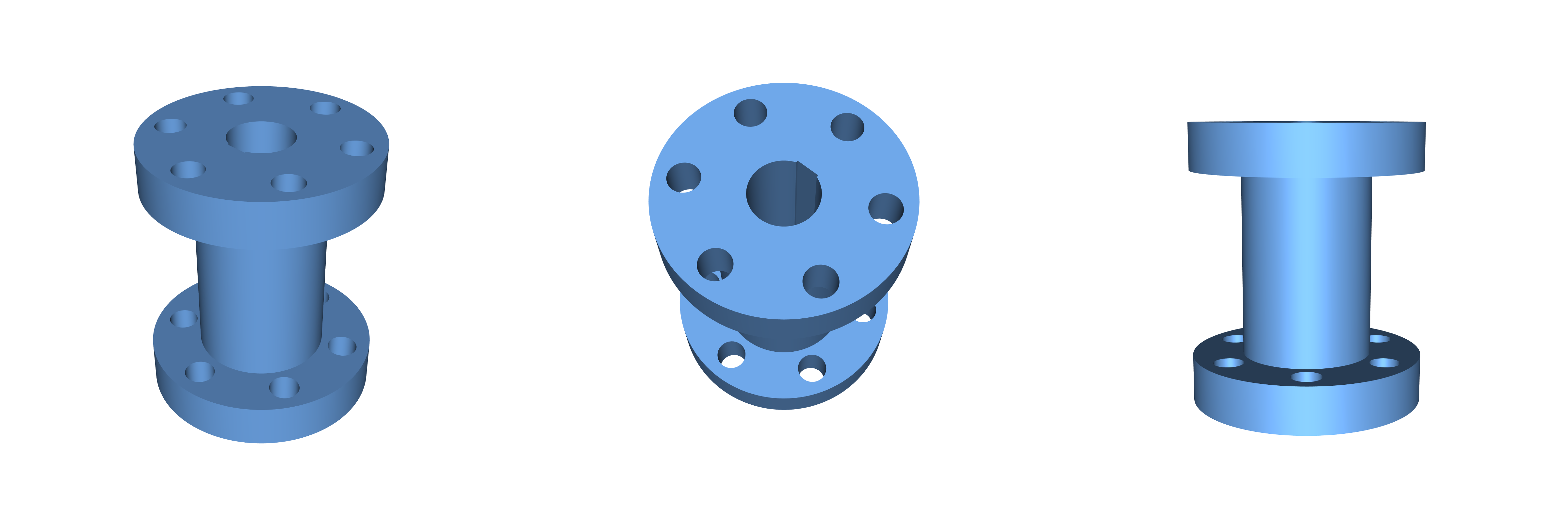}
    \caption{Three-view render provided to the Validator Judge at each iteration. Left: isometric view showing overall shape. Center: high-angle rear view revealing top-face features (bolt holes, center bore). Right: front profile showing the vertical structure (flanges, hub, keyway). These views are rendered from the generated STL using VTK with Phong shading. The Judge cross-references what it sees in these views against the kernel metrics and the original prompt to catch failures that numerical checks alone would miss.}
    \label{fig:vision_views}
\end{figure*}

The Judge receives four inputs: (1) the original user prompt, (2) the generated CadQuery code, (3) exact kernel measurements from the Executor (volume, bounding box, face/edge/vertex counts, center of mass, solid validity), and (4) a three-view rendered image of the generated part. The three views, shown in Fig.~\ref{fig:vision_views}, are rendered using VTK with Phong shading at 2400$\times$800 resolution:

\begin{itemize}[nosep,leftmargin=*]
    \item \textbf{Isometric} (elevation 35\textdegree, azimuth 45\textdegree): overall 3D shape.
    \item \textbf{High-angle rear} (elevation 65\textdegree, azimuth 220\textdegree): top-face features such as holes, bores, and cavities.
    \item \textbf{Front profile} (elevation 10\textdegree, azimuth 0\textdegree): vertical profile, wall heights, and layered features.
\end{itemize}

The Judge cross-references visual evidence with kernel metrics and the code to produce a pass/fail decision with analytical feedback. It is instructed to count features (holes, mounting points) visible in the renders and verify counts against the prompt, check that bounding box dimensions align with specified values, and confirm that all requested features are present in both the code and the rendered views. A hard kernel check is also enforced deterministically: if the solid is not valid (not watertight), the iteration fails regardless of the Judge's assessment.

The Judge also receives its own prior feedback from previous iterations. If the same issue persists across iterations, the Judge escalates by recommending a fundamentally different construction approach rather than repeating the same suggestion.

The Validator Judge prompt, which governs this evaluation, is shown in Fig.~\ref{fig:judge_prompt}.

\begin{figure}[t]
\begin{tcolorbox}[
    colback=blue!3,
    colframe=blue!40,
    fonttitle=\bfseries\small,
    title={Validator Judge System Prompt (condensed)},
    boxrule=0.4pt,
    arc=2pt,
    left=4pt, right=4pt, top=2pt, bottom=2pt,
    fontupper=\scriptsize
]
You are the Validator Agent. You receive: (1) the original prompt, (2) the CadQuery code, (3) OCCT kernel metrics, (4) a three-view rendered image.

\textbf{Evaluation rules:}
\begin{itemize}[nosep,leftmargin=*,label=\textbullet]
    \item Cross-reference ALL evidence: code logic, kernel measurements, and rendered views
    \item Count holes/bolts/mounting points in the rendered views and verify against the prompt
    \item Check bounding box dimensions against the prompt requirements
    \item Catch false convergence: metrics may look acceptable while the shape is fundamentally wrong
\end{itemize}

\textbf{Escalation:} If you gave feedback before and the same issue persists, do not repeat the same suggestion. Recommend a fundamentally different construction approach.

\textbf{Output:} \texttt{\{``passed'': bool, ``feedback'': ``...''\}}
\end{tcolorbox}
\caption*{}
\vspace{-12pt}
\captionof{figure}{Condensed Validator Judge system prompt. The full prompt includes detailed instructions for interpreting each of the three rendered views, handling non-rectangular bounding boxes, and formatting feedback for the Refiner agent.}
\label{fig:judge_prompt}
\end{figure}

\subsection{Geometric Refinement}

When the Validator fails a part, the Refiner agent receives the current code, the Judge's feedback (with exact numeric discrepancies), the design plan, and the original prompt. It produces a corrected version of the code targeting the specific failures identified. The Refiner also receives the full history of previous refinement attempts, including what feedback was given and what changes were tried, to prevent oscillation. At iteration three and beyond, the Refiner receives explicit escalation guidance to reconsider the overall construction strategy rather than continuing to adjust the same parameters.

The corrected code is sent back to the Executor, restarting the outer loop. This continues until the Validator passes the geometry or the maximum of five refinement iterations is reached.

\subsection{Evaluation Metrics}

We adopt the same three metrics used by Text-to-CadQuery \cite{xie2025texttocadquery} to enable direct comparison:

\textbf{Chamfer Distance (CD).} The average bidirectional squared nearest-neighbor distance between 10{,}000 surface points sampled from each mesh. Lower values indicate closer surface alignment.

\textbf{F1 Score.} The harmonic mean of precision and recall at a distance threshold $\tau$. A sampled point is ``correct'' if its nearest neighbor on the other surface is within $\tau$. We use $\tau = 1.0$\,mm in absolute space.

\textbf{Volumetric IoU.} The intersection-over-union of voxelized occupancy grids. Both meshes are voxelized at 1.0\,mm resolution, with adaptive coarsening for parts exceeding 100\,mm extent to prevent memory exhaustion.

All metrics are computed in absolute millimeter space rather than normalized coordinates. This is a deliberate departure from the normalized [0,1]\textsuperscript{3} evaluation used in prior work \cite{xie2025texttocadquery, wu2021deepcad}: normalized metrics erase dimensional accuracy, scoring a 10\,mm box and a 100\,mm box identically if their shapes match. Since our benchmark specifies explicit millimeter dimensions, absolute-space comparison captures both shape and dimensional accuracy.

Before computing metrics, both meshes are co-registered by translating their bounding box centers to the origin, then aligned using Iterative Closest Point (ICP) to resolve orientation mismatches from different construction frames.

\subsection{Ablation Configurations}

The system described above represents the full CADSmith pipeline. To isolate the contributions of each component, we evaluate two additional configurations:

\textbf{Zero-shot baseline.} A single Claude Sonnet call with a minimal system prompt. No Planner, no RAG, no refinement loop, no Validator, and no vision. The model receives only the raw prompt and must produce CadQuery code in one pass.

\textbf{No-vision ablation.} The full pipeline (Planner, Coder with RAG, Executor, Validator Judge, Refiner) with the three-view rendered image removed from the Judge's input. The Judge still receives the prompt, code, and kernel metrics, but makes its assessment without visual evidence.

\section{Results}

We evaluate CADSmith on all 100 benchmark entries across three configurations: the full pipeline with vision, the full pipeline without vision, and a zero-shot baseline. All metrics are computed in absolute millimeter space with ICP alignment.

\subsection{Overall Performance}

Table~\ref{tab:overall} compares the three configurations. The full pipeline achieves 100\% execution success and the highest scores across all metrics. The difference between configurations is most visible in mean Chamfer Distance, which captures outlier failures that the median obscures: the zero-shot baseline averages 28.37 due to a small number of catastrophic mismatches, while the full pipeline reduces this to 0.74.

\begin{table}[!ht]
\caption{Overall results across 100 benchmark entries.}
\label{tab:overall}
\centering
\small
\begin{tabular}{l c c c c c}
\hline
\textbf{Configuration} & \textbf{Exec\%} & \textbf{CD\textsubscript{med}} & \textbf{CD\textsubscript{mean}} & \textbf{F1\textsubscript{med}} & \textbf{IoU\textsubscript{med}} \\
\hline
Zero-shot       & 95  & 0.55  & 28.37 & 0.9707 & 0.8085 \\
No-vision       & 99  & 0.48  & 18.19 & 0.9792 & 0.9563 \\
Full (vision)   & 100 & 0.48  & 0.74  & 0.9846 & 0.9629 \\
\hline
\end{tabular}
\end{table}

Of the 100 entries in the full pipeline, 88 converged on the first iteration (iteration~0), with an average of 0.13 refinement iterations per entry. The remaining 12 entries required one or two additional iterations before the Validator accepted the geometry.

\subsection{Per-Tier Breakdown}

Table~\ref{tab:pertier} shows the full pipeline results by difficulty tier. Performance degrades smoothly from T1 to T3, reflecting the increasing geometric complexity. T1 and T2 achieve near-perfect F1 scores (0.999 and 0.998), while T3 parts average 0.886, consistent with the multi-step construction and feature interactions these parts require. The impact of vision is most pronounced on T3, where removing the rendered image from the Judge increases mean Chamfer Distance from 1.42 to 49.68 and drops mean F1 from 0.85 to 0.74. On T1 and T2, the no-vision configuration performs comparably, indicating that kernel metrics alone are sufficient for simpler parts but not for complex geometry where false convergence is more likely.

\begin{table}[!ht]
\caption{Full pipeline (vision) results by difficulty tier.}
\label{tab:pertier}
\centering
\small
\begin{tabular}{l c c c c c}
\hline
\textbf{Tier} & \textbf{N} & \textbf{CD\textsubscript{med}} & \textbf{CD\textsubscript{mean}} & \textbf{F1\textsubscript{med}} & \textbf{IoU\textsubscript{med}} \\
\hline
T1 (Primitives)   & 50 & 0.32 & 0.47 & 0.9985 & 0.9834 \\
T2 (Engineering)  & 25 & 0.32 & 0.58 & 0.9979 & 0.7661 \\
T3 (Complex)      & 25 & 0.96 & 1.42 & 0.8859 & 0.9582 \\
\hline
\end{tabular}
\end{table}

\subsection{Largest Improvements Over Zero-Shot}

The full pipeline produces the largest gains on entries where the zero-shot baseline fails substantially. T3\_023 improved from F1\,=\,0.037 (zero-shot) to F1\,=\,0.943 (full pipeline), T1\_021 from 0.168 to 0.995, and T2\_005 from 0.095 to 0.867. In each case, the zero-shot model produced code that executed but generated fundamentally wrong geometry, which the refinement loop corrected through iterative feedback.

\subsection{Failure Modes}

While the full pipeline achieves strong aggregate results, individual failures remain. The bottom T3 entries (T3\_016 at F1\,=\,0.57, T3\_024 at F1\,=\,0.59) involve parts where the generated construction approach diverges significantly from the reference, producing geometrically valid but structurally different solids.

\begin{figure}[H]
    \centering
    \includegraphics[width=0.95\columnwidth]{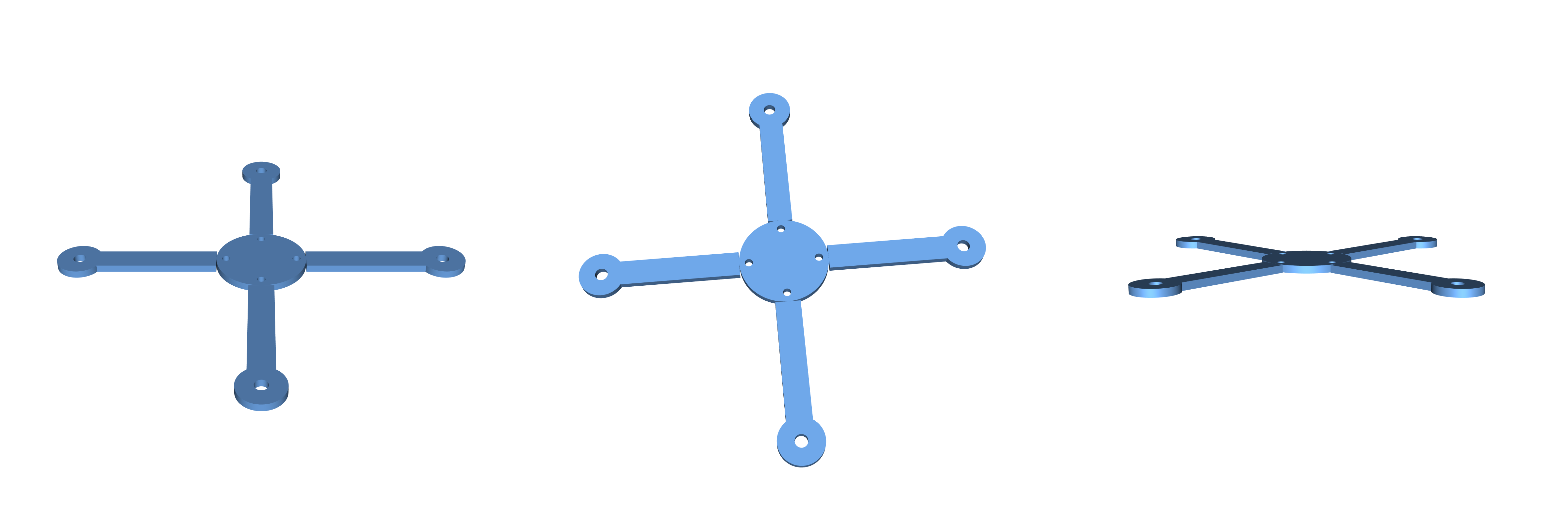}
    \caption{T3\_019 (quadcopter frame): F1\,=\,0.963, IoU\,=\,0.985. The part passed all validation checks, but contains subtle gaps between the arms and central hub (visible in the front profile view at right). This near-miss failure evaded both kernel metrics and the vision Judge.}
    \label{fig:drone_error}
\end{figure}

A subtler failure is illustrated in Fig.~\ref{fig:drone_error}. T3\_019 (quadcopter frame) scored F1\,=\,0.963 and IoU\,=\,0.985, converging on the first iteration. Both the metrics and the Validator Judge assessed this as correct. However, the generated part contains small gaps between the arms and the central hub that are difficult to detect in the rendered views. This represents a class of near-miss failures where the geometry is close enough to satisfy both programmatic checks and visual inspection, but would not be suitable for manufacturing. We return to this in Section~V.

\section{Discussion}

CADSmith demonstrates that a multi-agent architecture with programmatic geometric feedback substantially outperforms single-pass generation for text-to-CAD tasks. The full pipeline achieves 100\% execution success and reduces mean Chamfer Distance by 38$\times$ compared to the zero-shot baseline, confirming that structured feedback from the OpenCASCADE kernel provides the precision needed to drive iterative correction. Most prior work in text-to-CAD generation evaluates on single primitives or simple sketch-and-extrude parts \cite{wu2021deepcad, khan2024text2cad, xie2025texttocadquery}. Our T3 tier shows that LLM-based generation can produce multi-feature engineering parts involving lofts, sweeps, shells, and multi-body unions, achieving a median F1 of 0.886 on these complex geometries. The ablation results highlight that vision is critical to this performance: removing the three-view rendered image from the Judge increases T3 mean Chamfer Distance from 1.42 to 49.68, a 35$\times$ degradation. Kernel metrics alone cannot detect false convergence, where a shape has plausible volume and bounding box dimensions but is structurally wrong. The vision Judge catches these failures by cross-referencing what it sees against the prompt and the code, providing a complementary feedback channel that purely numerical checks cannot replace. At a time when there is renewed emphasis on bringing manufacturing capability back to the United States, systems that accelerate CAD workflows and lower the barrier to producing valid engineering geometry have practical significance beyond the research contribution.

However, the T3\_019 failure (Fig.~\ref{fig:drone_error}) reveals a limitation of the current vision approach. Three fixed isometric views are not always sufficient to detect small-scale geometric errors such as gaps at joints or incomplete boolean unions. The vision Judge assessed the quadcopter frame as correct because the gaps between arms and hub are difficult to resolve at the rendered scale and viewing angles. Future work could address this through adaptive view selection that targets areas of geometric complexity, higher-resolution crops of joints and interfaces, or additional views driven by the Judge's own uncertainty about specific features. Beyond vision improvements, natural extensions include evaluation across multiple frontier models, support for multi-part assemblies, and scaling the benchmark to capture a broader range of manufacturing-relevant geometries. CADSmith establishes that closed-loop refinement with combined programmatic and visual feedback is a viable path toward reliable automated CAD generation from natural language.

\bibliographystyle{IEEEtran}
\bibliography{references}

\end{document}